\documentclass[letterpaper, 10 pt, conference]{ieeeconf}  
\IEEEoverridecommandlockouts                              
\usepackage[noadjust]{cite}
\usepackage{graphicx}
\usepackage{amsmath}
\usepackage{algpseudocode}
\usepackage{algorithm}
\usepackage[algo2e]{algorithm2e}
\usepackage{multirow}
\usepackage[normalem]{ulem}
\usepackage{hyperref}

\DeclareMathOperator*{\argmax}{arg\,max}
\hypersetup{
    colorlinks=true,
    linkcolor=blue,
    filecolor=magenta,      
    urlcolor=blue,
}

\usepackage{verbatim}

\title{\LARGE \bf CASSL: Curriculum Accelerated Self-Supervised Learning}
\author{%
Adithyavairavan Murali, Lerrel Pinto, Dhiraj Gandhi, Abhinav Gupta \\
The Robotics Institute, Carnegie Mellon University \\
\texttt{\{amurali, lerrelp, dgandhi, gabhinav\}@andrew.cmu.edu}
}%

\begin{document}
\maketitle
\thispagestyle{empty}
\pagestyle{empty}

\begin{abstract}
Recent self-supervised learning approaches focus on using a few thousand data points to learn policies for high-level, low-dimensional action spaces. However, scaling this framework for higher-dimensional control requires either scaling up the data collection efforts or using a clever sampling strategy for training. We present a novel approach - Curriculum Accelerated Self-Supervised Learning (CASSL) - to train policies that map visual information to high-level, higher-dimensional action spaces. CASSL orders the sampling of training data based on control dimensions: the learning and sampling are focused on few control parameters before other parameters. The right curriculum for learning is suggested by variance-based global sensitivity analysis of the control space. We apply our CASSL framework to learning how to grasp using an adaptive, underactuated multi-fingered gripper, a challenging system to control. Our experimental results indicate that CASSL provides significant improvement and generalization compared to baseline methods such as staged curriculum learning (8\% increase) and complete end-to-end learning with random exploration (14\% improvement) tested on a set of novel objects.
\end{abstract}

\section{Introduction}
With the advent of big data in robotics~\cite{pinto2016supersizing,levine2016learning,agarwal2016, levine2016end}, there has been an increasing interest in self-supervised learning for planning and control. The core idea behind these approaches is to collect large-scale datasets where each data-point has the current state (e.g. image of the environment), action/motor-command applied, and the outcome (success/failure/reward) of the action. This large-scale dataset is then used to learn policies, typically parameterized by high-capacity functions such as Convolutional Neural Networks (CNNs) that predict the actions of the agent from input images/observations. But what is the right way to collect this dataset for self-supervised learning?

Most self-supervised learning approaches use random exploration. That is, first a set of random objects is placed on the table-top followed by a random selection of actions. However, is random sampling the right manner for training a self-supervised system? Random exploration with few thousand data points will only work when the output action space is low-dimensional. In fact, the recent successes in self-supervised learning which shown experiments on real robots (not just simulation) use a search space of only 3-6 dimensions~\footnote{\cite{pinto2016supersizing,levine2016learning,agarwal2016} use 3,4,5-dim search space respectively} for output action space. Random exploration is also sub-optimal since it leads to a very sparse sampling of the action-space.

\begin{figure}[t!]
\label{fig:intro}
\centering
  \includegraphics[width = 3.0in ]{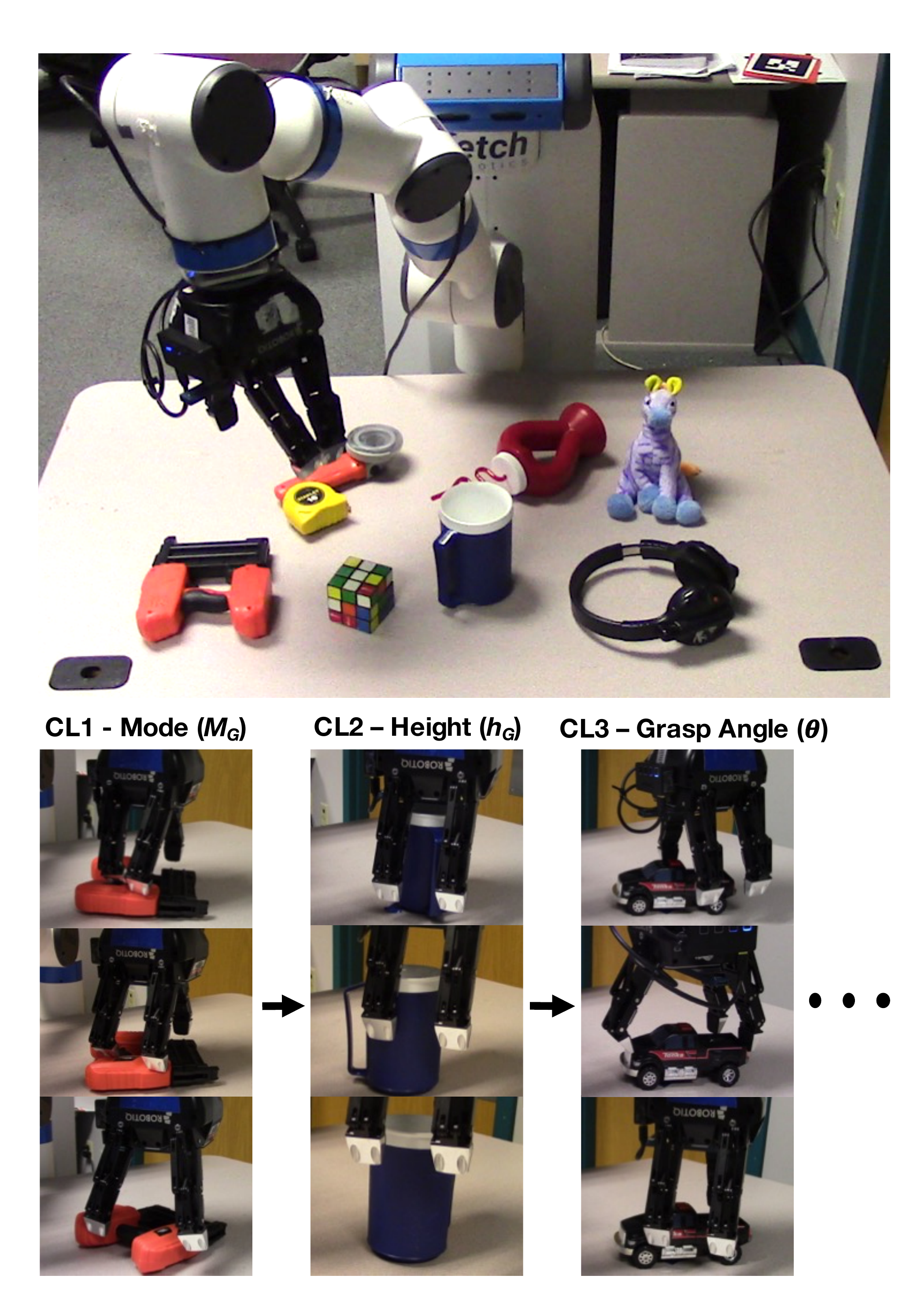}
  \caption{Given a table-top scene, our robot learns to grasp objects by Curriculum Accelerated Self-Supervised Learning (CASSL). Given the various control dimensions, such as mode, height, grasp angle, etc., our robot focuses on learning to predict the easier dimensions earlier. We used a Fetch-robot with an adaptive 3-fingered gripper from Robotiq.}
\end{figure}

In this paper, we focus on the problem of sampling and self-supervised learning for high-level, high-dimensional control. One possible approach is to collect and sample training data using staged-training~\cite{pinto2016supersizing} or on-policy search~\cite{sutton1996generalization}. In both these approaches, random sampling is first used to train an initial policy. This policy is then used to sample the next set of training points for learning. However, such approaches are hugely biased due to initial learning from random samples and often sample points from a small search space. Therefore, recent papers have investigated other exploration strategies, such as curiosity-driven exploration~\cite{houthooft20015variational}. However, data sparsity in high-dimensional action space still remains a concern.

Let's take a step back and think how do humans deal with high-dimensional control. We note that the action space of human control grows continually with experience: the search does not start in high-dimensions but rather in a small slice of the high-dimensional space. For example, in the early stages of human development, when hand-eye coordination is learned, a single mode of grasping (palmar-grasp) is used, and we gradually acquire more complex, multi-fingered grasping modalities~\cite{yasuyuki2012palmar}. Inspired by this observation, we propose a similar strategy: order the learning in control parameter space by fixing few dimensions of control parameters and sampling in the remaining dimensions. We call this strategy curriculum learning in control space, where the curriculum decides which control dimensions to learn first~\footnote{Note our curriculum is defined in control space as opposed to standard usage where easy examples are used first followed by hard examples for training. In our case, the objects being explored, though diverse and numerous, remain fixed.}. We use a sensitivity analysis based approach to define the curriculum over control dimensions. We note that our framework is designed to infer high-level control commands and use planners/low-level controllers to achieve desired commands. In future work, the curriculum learning of low-level control primitives, such as actuator torques, could be explored.

We demonstrate the effectiveness of our approach for the task of adaptive multi-fingered grasping (See Fig~\ref{fig:intro}). Our search space is 8-dimensional and we sample the training points for learning control in 6-dimensions ($x,y$ is done via region-proposal sampling, as explained later). We show how a robust model for grasping can be learnt with very few examples. Specifically, we illustrate that defining a curriculum over the control space improves overall grasping rate compared to that of random sampling and staged-training strategy by a significant margin. To the best of our knowledge, this is the first work applications of curriculum learning on a physical robotic task.

\section{Related Work}
{\bf Curriculum Learning:} For biological agents, concepts are easier to learn when provided in a structured manner instead of an arbitrary order~\cite{skinner1958}. This idea has been formalized for machine learning algorithms by Elman et al.~\cite{elman1993} and Bengio et al.~\cite{bengio2009curriculum}. Under the name of Curriculum Learning (CL) ~\cite{bengio2009curriculum}, the core idea is to learn easier aspects of the problem earlier while gradually increasing the difficulty. Most curriculum learning frameworks focus on the ordering of training data: first train the model on easy examples and then on more complex data points. Curriculum over data has been shown to improve generalization and speed up convergence~\cite{ffl2010,xinlei2015}. In our work, we propose curriculum learning over the control space for robotic tasks. The key idea in our method is that in higher dimensional control spaces, some modalities are easier to learn and are uncorrelated with other modalities. Our variance-based sensitivity analysis exposes these easy to learn modalities which are learnt earlier while focusing on harder modalities later.

{\bf Intrinsic Motivation:} Given the challenges for reinforcement learning in tasks with sparse extrinsic reward, there have been several works that have utilized intrinsic motivation for exploration and learning. Recently, Pathak et. al. learned a policy for a challenging visual-navigation task by optimizing with intrinsic rewards extracted from self-supervized future image/state prediction error \cite{pathak2017curiosity}. Sukhbaatar et al. proposed a asymmetric self-play scheme between two agents to improve data efficiency and incremental exploration of the environment \cite{selfplay2017}. In our work, the curriculum is defined over the control space to incrementally explore parts of the high-dimensional action space.

{\bf Ranking Functions:} An essential challenge in CL is to construct a ranking function, which assigns the priority for each training datapoint. In situations with human experts, a stationary ranking function can be hand defined. In Bengio et al.~\cite{bengio2009curriculum}, the ranking function is specified by the variability in object shape. Some other methods like Self-Paced Learning \cite{koller2010} and Self-Paced Curriculum Learning \cite{hauptmann2015} dynamically update the curriculum based on how well the agent is performing. In our method, we use a stationary ranking that is learned from performing sensitivity analysis~\cite{saltelli2010variance} on some data collected by sampling the control values from a quasi-random sequence. This stationary ranking gives priority ordering on control parameters. Most formulations of curriculum training use a linear curriculum ordering. A recent work by Svetlik et al. generated a directed acyclic graph of curriculum ordering and showed improved data efficiency for training an agent to play Atari games with reinforcement learning \cite{Svetlik2017}.

{\bf Grasping:} We demonstrate data-efficiency of CASSL on the grasping problem. Refer to \cite{bicchi2000robotic, bohg2014data} for surveys of prior work. Classical foundational approaches focus on physics-based analysis of stability~\cite{nguyen1988constructing}. However, these methods usually require explicit 3D models of the objects and do not generalize well to unseen objects. To perform grasping in complex unstructured environments, several data-driven methods have been proposed~\cite{mahler2016dexnet,pinto2016supersizing,levine2016learning}. For large-scale data collection both simulation~\cite{mahler2016dexnet} and real-world robots~\cite{pinto2016supersizing,levine2016learning} have been used.
However, these large scale methods operate on lower dimensional control spaces (planar grasps are often 3 dimensional in output space) since high-dimensional grasping requires significantly more amount of data. In our work, we hypothesize and show that CASSL requires lesser data and can also learn on higher dimensional grasping configurations.

{\bf Robot Learning:} The proposed method of Curriculum Accelerated Self-Supervised Learning~(CASSL) is not specific to the task of grasping and can be applied to a wide variety of robot learning, manipulation and self-supervised learning tasks. The ideas of self-supervised learning have been used to push and poke objects ~\cite{agarwal2016,pinto2016mlt}. Nevertheless, a common criticism of self-supervised approaches is their dependency on large scale data. While reducing the amount of data for training is an active area for research~\cite{pinto2016supervision}, CASSL may help in reducing this data dependency. Deep reinforcement learning~\cite{mnih2015human,schulman2015trust,duan2016benchmarking} methods have empirically shown the ability of neural networks to learn complex policies and general agents. Unfortunately, these model-free methods often need data in the order of millions to learn their perception-based control policies. 


\section{Curriculum Accelerated Self-Supervised Learning (CASSL)}
We now describe our curriculum learning approach for high-level control. First, we discuss how to obtain priority ordering of control parameters followed by how to use the curriculum for learning.

\subsection{CASSL Framework}
Our goal is to learn a policy $v=\pi(I)$ and scoring function $y=\mathcal{F}(I, v)$, which given the current state represented by image $I$ and action $v$ predicts the likelihood of success $y$ for the task. Note that in the case of high-dimensional control $v={v_1, v_2....v_K}$ where $K$ is the dimensionality of the action space. For the task of grasping an object, $y$ can be the grasp success probability given the image of object ($I$) and control parameters of the grasp configuration ($v$). The high-level control dimensions for grasping are the grasping configuration, gripper pose, force, grasping mode, etc. as explained later.

The core idea is that instead of randomly sampling the training points in the original K-dim space and learning a policy, we want to focus learning on specific dimensions first. So, we will sample more uniformly~(high exploration) in the dimensions we are trying to learn; and for the other dimensions we use the current model predictions~(low exploration). Consequently, the problem is reduced to the challenge of finding the right ordering of the control dimensions. One way of determining this ranking is with expert human labeling. However, for the tasks we care about, the output function $\mathcal{F}(I,v)$ is often too complex for a human to infer rankings due to the complex space of grasping solutions. Instead, we use global sensitivity analysis on a dataset of physical robotic grasping interactions to determine this ranking. The key intuition is to sequentially select the dimension that is the most independent and interacts the least with all other dimensions, hence is easier to learn.

\begin{figure*}
\centering
\includegraphics[width = 0.95\linewidth]{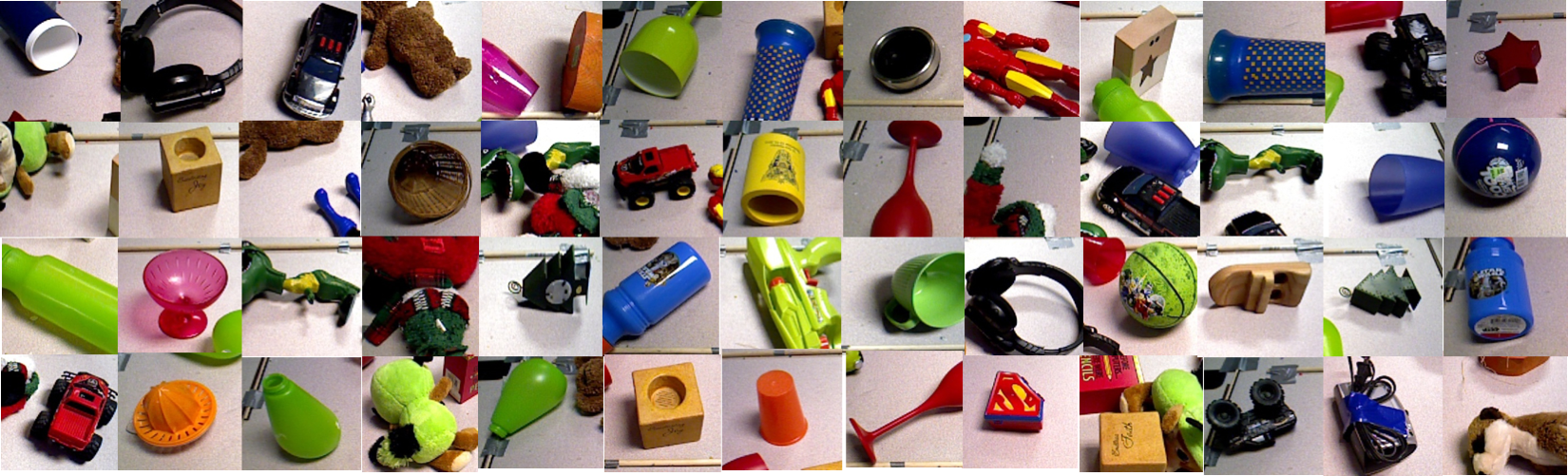}
  \vspace{-0.1in}
\caption{A small subset of the data processed by the model during training can be seen here. Note that during training, we use a wide variety of objects with different sizes, shapes and rigidity. 
}
  \vspace{-0.1in}
\label{fg:collage}
\end{figure*}

\subsection{Sensitivity Analysis} 
\label{sc:sensitivity_analysis}
For defining a curriculum over control dimensions, we use variance-based global sensitivity analysis. Mathematically, for a model of the form $y = \mathcal{F}(I,v=\{v_{1},v_{2}, \cdots, v_{K}\})$, global sensitivity analysis aims to numerically quantify how uncertainty in the scalar output (e.g. grasp success probability in this paper) can be expressed in terms of uncertainty in the input variables (i.e. the control dimensions)~\cite{sensitivity2010}. The first order index, denoted by $S_{j}^{(1)}$, is the most preliminary metric of sensitivity and represents the uncertainty in $y$ that comes from $v_j$ alone. Another metric of interest is the total sensitivity index $S_{j}^{(T)}$, which is the sum of all sensitivity indices (first and higher order terms) involving the variable $v_j$. As a result, it captures the interactions (pairwise, tertiary, etc.) of $v_j$ with other variables. Detailed description on monte carlo estimators for the indices and proofs can be found in~\cite{sensitivity2010}. Obtaining the sensitivity metrics requires the model $\mathcal{F}$ or an approximate version of it. Instead, we use Sobol sensitivity analysis~\cite{salib2017} implementation in \texttt{SAlib} and propose a data-driven method for estimating the sensitivity metrics. In Sobol sensitivity analysis, the control input is sampled from a quasi-random sequence, as it provides a better coverage/exploration of the control space compared to a uniform random distribution.

\subsection{Determining the Curriculum Ranking}
Given a large control space, an intuitive curriculum would be to learn control dimensions in the \textit{descending} order of their sensitivity. However, when designing a curriculum, we also care about the interactions between a control dimension and others. Hence, we need to optimize on getting dimensions that have high sensitivity and low correlation with other dimensions. One way to do this is to minimize higher order ($>$1) terms (i.e. $S_{i}^{(T)} - S_{i}^{(1)}$) and the pairwise interactions between variables $S_{i}^{(2)}$. Given sensitivity values for each control dimension, we choose the subset of dimensions $\Psi$ which minimize the heuristic Eqn~\ref{eq:optim} below:

\begin{equation}
\min_{\Psi} E(\Psi) = \sum_{i \in \Psi} (S_{i}^{(T)} - S_{i}^{(1)}) + \sum_{i \in \Psi} \sum_{j \in (\Omega - \Psi)} \lvert S_{ij}^{(2)} \rvert
\label{eq:optim}
\end{equation}

Here $\Omega$ is the set of all control dimensions (i.e. $\Omega$ = $\{v_{1},v_{2}, \cdots, v_{K}\}$), and $\Psi$ is a subset of dimensions. We evaluate all possible $2^K-1$ subsets and choose the subset with the minimum value as the first set of control dimensions in the curriculum. We then recompute the term for subsets of remaining control dimensions and iteratively choose the next subset (as seen in Algorithm~\ref{alg:curr_train}). The intuition behind Eqn~\ref{eq:optim} is that we want to choose the subset of control dimensions on which the output $y$ depends the most and is least correlated with the remaining dimensions.

\begin{figure*}
\centering
\includegraphics[width = 0.95\linewidth]{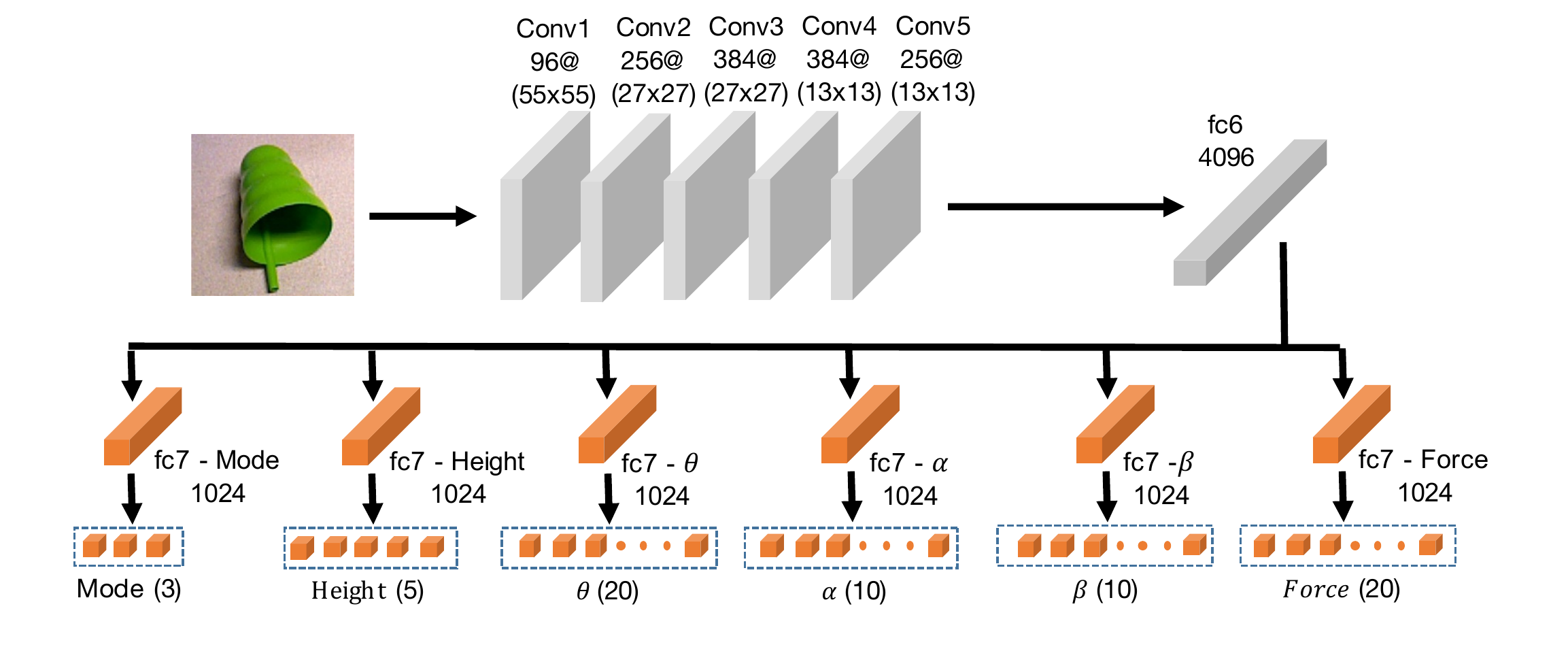}
\vspace{-0.2in}
\caption{We employ a deep neural network to learn the action policy. The convolutional layers and the first fully connected layer (fc6) are shared (in grey). The fc7 and output control layers are trained (in orange) to learn control-specific weights.
}
\vspace{-0.17in}
\label{fig:architecture}
\end{figure*}

\subsection{Modeling the Policy}
\label{sc:policy}
The policy function $v =\pi(I)$ takes the image as input $I$ and outputs the desired action $v$. Inspired by the approach in ~\cite{pinto2016supersizing}, we use a CNN to model the policy function.  However, since CNNs have been shown to work better on classification than regression, we employ classification instead of regressing control outputs. To this end, each control space is discretized into $n_i$ bins as given in Table \ref{tab:discretization}. 

Our network design is based on AlexNet~\cite{krizhevsky2012imagenet}, where the convolutional layers are initialized with ImageNet~\cite{olga2014imagenet} pre-trained weights as done before in \cite{pinto2016supersizing, pinto2016curious}. We used ImageNet pre-trained features as they been proven to be effective for transfer learning in a number of visual recognition tasks \cite{ross2014, razavian2014}. The network architecture is shown in Fig~
\ref{fig:architecture}. The fully-connected layer's weights are initialized from a normal distribution. While we could have had separate networks for each control parameter, this would enormously increase the size of our model and make the prediction completely independent. Instead, we employ a shared architecture commonly used in multi-task learning~\cite{pinto2016curious,pinto2016mlt}, such that the non-linear relationship between the different parameters could be learned. Each parameter has a separate fc7 layer and this ensures that the network learns a shared representation of the task until the fc6 layer. The fc8 ouputs are finally sent through and normalized by a sigmoidal function. Predicting the correct discretized value for each control parameter is formulated as a multi-way classification problem. More specifically, $p_{ij}=\pi(I, u_{ij})$ is akin to a Q value function that returns the probability of success when the action corresponding to the $j^{th}$ discrete bin for control dimension $i$ is taken.

\subsection{Curriculum Training}
Algorithm~\ref{alg:curr_train} describes the complete training structure of our method. First, initial data is collected to perform sensitivity analysis and given this priority ordering, we begin the training procedure for our policy models. Apart from diversity in the objects seen, we still need to enforce exploration in the action space through all stages of the curriculum training.

As described in Algorithm~\ref{alg:curr_train}, the greedy action corresponds to executing whatever control values the network predicts. The hyper-parameters, $\epsilon_{post}$ = 0.15 and $\epsilon_{pre}$ = 0.7, determine the probability of choosing a random action vis-\`a-vis the greedy one given by the policy. Therefore, for the control dimensions already learned, we are more likely to select the policy via the network. In our framework, for parameters that have already been learned in the curriculum (i.e $i < k$), they will have little exploration. In contrast, for control parameters with $i > k$, they have a great deal of exploration so that the data collected captures the higher order effects between control parameters. When $i = k$, the control is chosen with importance sampling explained as follows. The grasping policy is parameterized as a multi-class classifier on a discretized action space. As a result, the output value $p_{ij}$ from the final sigmoid layer for the $j^{th}$ discrete bin for control $i$ can be treated as a bernoulli random variable with probability $p_{ij}$. Here, the control value $u_{i}$ that is selected is the one which the model is most uncertain about and hence has the highest variance i.e $u_{i}$ = $\argmax_{j}$ $p_{ij}$(1- $p_{ij}$)). Taking the analytic derivative, the uncertainity is maximized when $p_{ij}$=0.5. This approach is similar to previous works such as \cite{pathak2017curiosity}, where actions were taken based on what the agent is most ``curious''/uncertain about and the curiosity reward is defined as the prediction error of the next state given the current state and action. Similarly, in \cite{houthooft20015variational}, the actions that maximize information gain about the agent's belief of the environment dynamics were taken.

\setlength{\textfloatsep}{0pt}
\begin{algorithm}[t!]
    \textbf{Given:} $\xi$, $\epsilon_{pre}$, $\epsilon_{post}$, $D=\{\}$ \\
    \textbf{Collect:} dataset $d_0$ with quasi-random control samples\\
    \textbf{Initialize:} aggregated dataset $D \leftarrow D \cup \{d_0\}$ \\
    {[$S^{(1)}, S^{(2)}, S^{(T)}$]} $\leftarrow$ \text{SensitivityAnalysis}($D$) \\
    \textbf{Find curriculum} $\mathcal{C}$ using [$S^{(1)}, S^{(2)}, S^{(T)}$] \\
    \textbf{Train:} Models $M^{0}_i$ with $D$ $\forall i$ \\
        \For{control (indexed by \textit{k}) in $\mathcal{C}$}
        {
            Collect new dataset $d_{k}$ with the policy below:
        
            \[ \pi_{CASSL} =  \begin{cases} 
                  \epsilon_{post}\text{-Greedy with } M^{k-1}_i & i < k \\
                  \text{Importance sampling of fc\_{8}} & i = k \\
                  \epsilon_{pre}\text{-Greedy with } M^{k-1}_i & i > k 
              \end{cases}
            \]\\
            
            Aggregate new dataset $D = \{D, d_{k}$\}\\
            Update Model $M_{k}$ with $D$
        }   
    \caption{Curriculum Accelerated Self-Supervised Learning (CASSL)}\label{alg:curr_train}
\end{algorithm}
\setlength{\textfloatsep}{5pt}

At each stage of the curriculum learning, we also aggregated the training dataset similar to DAgger~\cite{dagger2010} and prior work~\cite{pinto2016supersizing}. On stage \textit{k} of the curriculum, the network was fine-tuned on $D_{k} = $\{$D_{k-1}$,  $d_{k}$\}, where $d_{k}$ is the dataset collected in the current stage of the curriculum. We sample $d_{k}$ 2.5 times more than $D_{k-1}$ to give more importance to new datapoints.

\section{CASSL for Grasping}
We now describe the implementation of \text{CASSL} for the task of grasping objects. The grasping experiments and data are collected on a Fetch mobile manipulator \cite{fetch2016}. Visual data is collected using a PrimeSense Carmine 1.09 short-range RGBD sensor and we use a 3-finger adaptive gripper from Robotiq. The Expanding Space Tree (ESTk) planner from MoveIt is used to generate collision-free trajectories and state estimation is hand-designed similar to prior work~\cite{pinto2016supersizing} - using background subtraction to detect newly placed objects on the table. We further use depth images to obtain an approximate value for the height of objects.

\begin{figure*}
  \begin{center}
    \includegraphics[width = \linewidth]{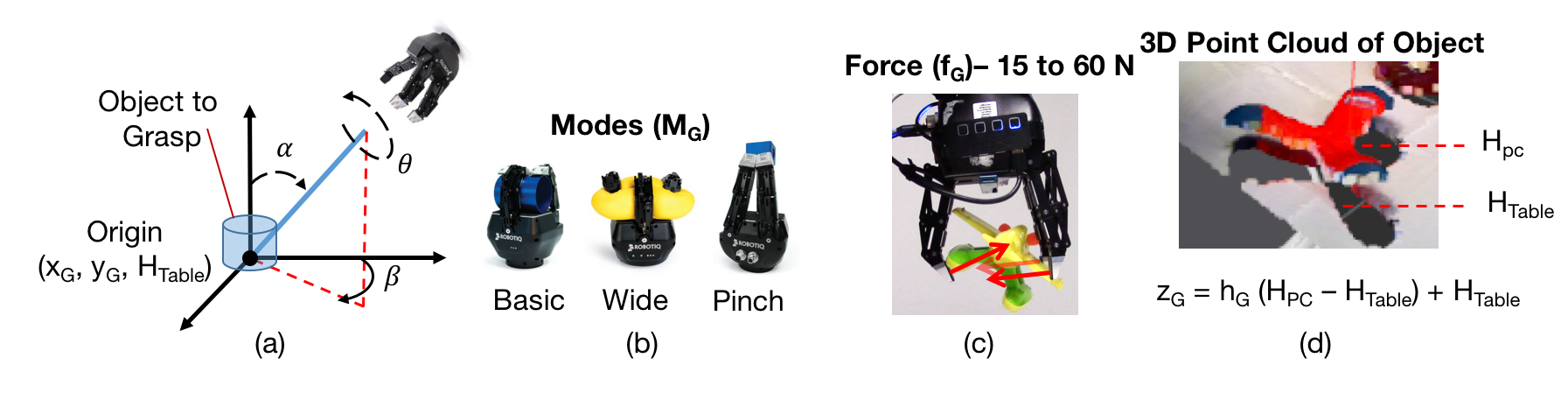}
  \end{center}
  \vspace{-0.25in}
  \caption{Our grasping problem formulation involves the high dimensional control of the adaptive gripper. (a) describes the translational and rotational control dimensions ($x_G$ and $y_G$ are however subsumed in input samples). (b) describes the various modes of grasping, i.e. basic, wide and pinch modes. (c) illustrates the force the gripper is allowed to apply on the objects. (d) describes the gripper's commanded height with respect to the table and the object.}
  \vspace{-0.1in}

\label{fg:problemdefinition}
\end{figure*}

\subsection{Adaptive Grasping}
The robotiq gripper has three fingers that can be independently controlled and has two primary grasp modalities - encompassing and finger-tip grips. As shown in Fig \ref{fg:problemdefinition}, there are three operational modes for the gripper - Pinch, Normal and Wide. Pinch mode is meant for precision grasping of small objects and is limited to finger-tip grasps. Normal grasping mode is the most versatile and can grasp a wide range of objects with encompassing and finger-tip grasps. Similarly, Wide mode is adept at grasping circular or large objects. While the fingers can be individually controlled, we only command the entire gripper to open/close, and the proprietary planner handles the lower-level control for the fingers. The fingers are operated at a speed of 110mm/sec.

The adaptive mechanisms of the gripper also allow it to better handle the uncertainty in the object's geometry and pose. As a result of the adaptive closing mechanism, some of the grasps end up being similar to push-grasps~\cite{pushgrasping2010iros}. The gripper fingers sweep the region containing the object, such that the object ends up being pushed inside the fingers regardless of its starting position. Sometimes, such grasps may not have force closure and the object could slip out of the gripper.

\subsection{Grasping Problem Definition}
We formulate our problem in the context of table-top grasping, where we infer high-level grasp control parameters based on the image of the object. There are three parameters that determine the location of the grasp ($x_G, y_G$ and $h_G$), three parameters that determine the approach direction and orientation of the gripper ($\alpha, \beta$ and $\theta$) and two others that involve the configuration (Mode $M_G$ and Force $f_G$). The geometric description of the three angles with respect to the object pose is shown in Fig \ref{fg:problemdefinition} and details of each parameter are provided in Table~\ref{tab:discretization}. $\theta$ is very sensitive to asymmetrical, elongated objects while $\alpha$ - the angle from the vertical axis - allows the gripper to tilt its approach direction to grasp the objects from the side. The camera's point cloud data gives a noisy estimate of the object height, denoted by $H_{pc}$. Let $H_{Table}$ be the height of the table with respect to the robot base. Then, $h_G$ is a scaling parameter (between 0 and 1) that interpolates between these two values, where the final height of the object is $z_{G} = h_G \cdot (H_{pc} - H_{Table}) + H_{Table}$. The height of a grasp is crucial in ensuring that the gripper moves low enough to make contact with the object in the first place. However, note that the error in the height depends on both $h_G$ and the noisy depth measurement from the camera. As shown in Fig \ref{fg:problemdefinition}, there were only three discrete modes for the gripper provided by the manufacturer.

Although the total space of grasp control is 8 dimensional, two of the translational controls ($x_G$ and $y_G$) are subsumed in the sampling. Given an input image of the entire scene $I_{S}$, 150 patches $I_P$ are sampled which correspond to the different values of $x_G$ and $y_G$. Though this increases the inference time (since we have to input multiple samples), it also massively decreases the search space as a lot of the scene $(\{x_G,y_G\}$ corresponding to the background) is empty. Hence, only 6 dimensions of control $\{h_G,\alpha, \beta, \theta, M_G, f_G \}$ are learned for our task of grasping.

\begin{table}
\centering
\caption{Control parameters, range and discretization}
\label{tab:discretization}
\begin{tabular}{|l|l|l|l|}
\hline
Parameter & Min     & Max     & \# of Discrete Bins\\ 
\hline
$\theta$     & $-180^\circ$ &  $180^\circ$ & 20             \\ \hline
$\alpha$     & $-10^\circ$     & $10^\circ$      & 10             \\ \hline
$\beta$      & $-30^\circ$     & $30^\circ$      & 10             \\ \hline
$h_G$ (Height)         & 0       & 1       & 5              \\ \hline
$M_G$ (Mode)     & 0       & 2       & 3              \\ \hline
$f_G$ (Force)     & 15N       & 60N     & 20             \\ \hline

\end{tabular}
\end{table}

\subsection{Sensitivity Analysis on Adaptive Grasping}
As described in Section~\ref{sc:sensitivity_analysis}, we collect a dataset of 1960 grasp interactions using the sobol quasi-random sampling scheme with an accuracy of 21\% during data collection. The results for the $S_{i}^{(1)}, S_{i}^{(T)}$ and $S_{ij}^{(2)}$ indices for all control parameters are shown in Table~\ref{tab:sens}. While the sensitivity analysis was limited to 10 objects, they were diverse in their properties - shape, deformable vs. rigid, large vs. small.  Given sensitivity indices for each control parameter, the objective function in Eqn~\ref{eq:optim} is optimized to determine the optimal ordering of the control parameters to learn. 
The ordering that minimizes Eqn~\ref{eq:optim} is: $[h_G, \theta, f_G, M_G, \alpha, \beta]$ in decreasing order of priority.

\begin{table}
\centering
\caption{Sensitivity Analysis results}

\label{tab:sens}
\begin{tabular}{|l|c|c|l|l|l|l|}
\hline
                            & \multicolumn{1}{l|}{$f_G$} & \multicolumn{1}{l|}{$M_G$}   & $\alpha$                  & $\beta$                   & $\theta$                  & $h_G$                      \\ \hline
$S^{(1)}$                          & \multicolumn{1}{l|}{0.014} & \multicolumn{1}{l|}{0.109}  & 0.040                  & 0.087                  & 0.164                  & 0.124                  \\ \hline
$S^{(T)}$                          & \multicolumn{1}{l|}{0.799} & 0.985                       & 0.892                  & 1.130                  & 0.850                  & 0.788                  \\ \hline
\multicolumn{7}{|c|}{$S^{(2)}$}                                                                                                                                                                   \\ \hline
\multicolumn{1}{|c|}{$f_G$} & -                          & \multicolumn{1}{l|}{0.0125} & -0.195                 & -0.216                 & -0.153                 & 0.0956                 \\ \hline
$M_G$                        & -                          & -                           & -0.0859                & 0.163                  & -0.190                 & 0.0385                 \\ \hline
$\alpha$                       & -                          & -                           & \multicolumn{1}{c|}{-} & -0.0904                & -0.194                 & -0.236                 \\ \hline
$\beta$                        & -                          & -                           & \multicolumn{1}{c|}{-} & \multicolumn{1}{c|}{-} & -0.280                 & -0.0519                \\ \hline
$\theta$                       & -                          & -                           & \multicolumn{1}{c|}{-} & \multicolumn{1}{c|}{-} & \multicolumn{1}{c|}{-} & -0.260                 \\ \hline
$h_G$                           & -                          & -                           & \multicolumn{1}{c|}{-} & \multicolumn{1}{c|}{-} & \multicolumn{1}{c|}{-} & \multicolumn{1}{c|}{-} \\ \hline
\end{tabular}
\end{table}

\subsection{Training and Model Inference}
Eqn~\ref{eq:loss} is the joint loss function that is optimized. $\hat{y}$ corresponds to the success/failure label, $D(i)$ gives the number of discretized bins for control parameter $i$ (see Table~\ref{tab:discretization}), K (=6) is the number of control parameters, B is the batch size and $\sigma$ is the sigmoid activation. $\delta(k, u_{i,j})$ is an indicator function and is equal to 1 when the control parameter $i$ corresponding to bin $j$ is applied. $y_{i,j}^{fc7}$ is the corresponding feature vector that is passed into the final sigmoid activation.

\begin{equation}
\footnotesize{
L = \sum_{i=1}^{K} \sum_{j=1}^{D(i)} \sum_{k=1}^{B} \delta(k, u_{i,j}) \cdot \text{Cross-Entropy}(\sigma(y_{i,j}^{fc7}),\hat{y}) 
\label{eq:loss}}
\end{equation}
Note that for each image datapoint, the gradients for all six control parameters are back-propagated throughout training. For each stage of the curriculum, the network is trained for 15-20 epochs with a learning rate of 0.0001 using the ADAM optimizer~\cite{kingma2014adam}. For inference, once we have the bounding box of the object of interest, 150 image patches are sampled randomly within this window and are re-sized to 224 $\times$ 224 dimensions for the forward pass through the CNN. For each control parameter, the discrete bin with the highest activation is selected and interpolated to obtain the actual continuous value. The networks and optimization are implemented in TensorFlow~\cite{abadi2016tensorflow}. As a good practice when training deep models, we used dropout(0.5) to reduce model over-fitting.

\section{Experimental Evaluation}
\begin{figure*}
  \begin{center}
    \includegraphics[width = 0.9\linewidth]{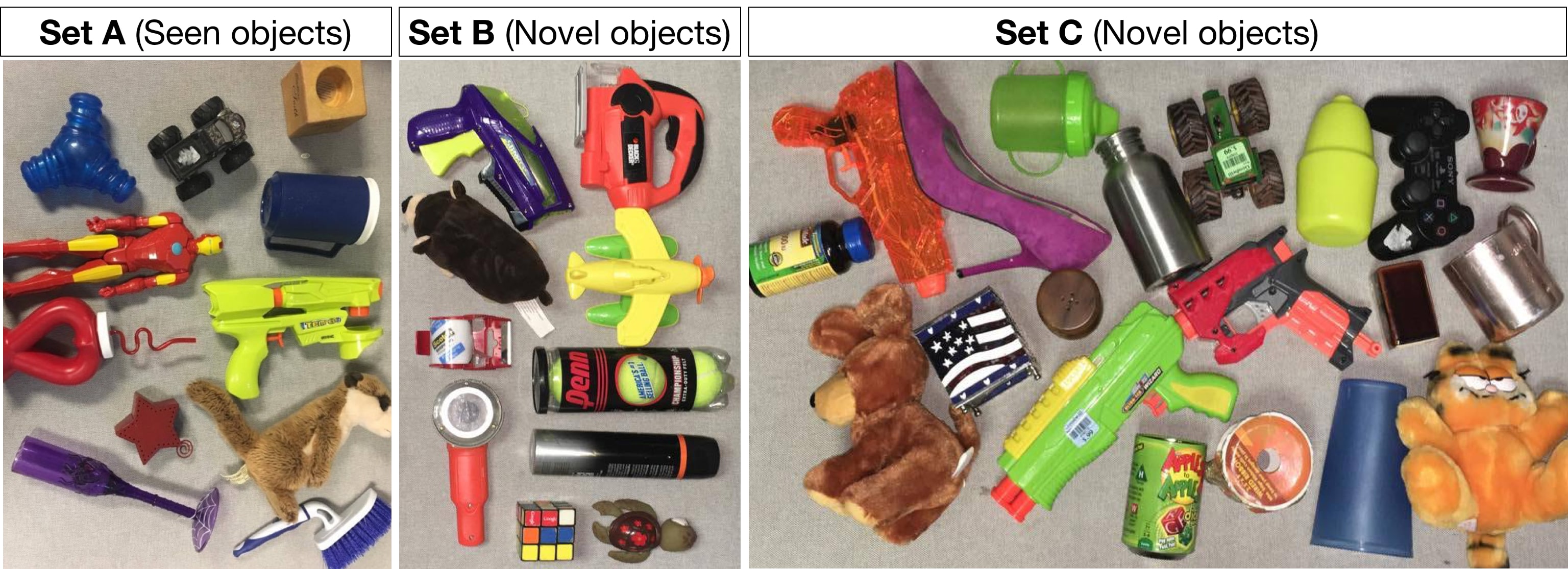}
  \end{center}
  \vspace{-0.2in}

  \caption{Set A contains 10 objects seen in training. Set B and C contain 10 and 20 novel objects respectively not used in training}
  \vspace{-0.1in}
  \label{fg:test_objects}
\end{figure*}

\begin{figure}
  \includegraphics[width =3.3in]{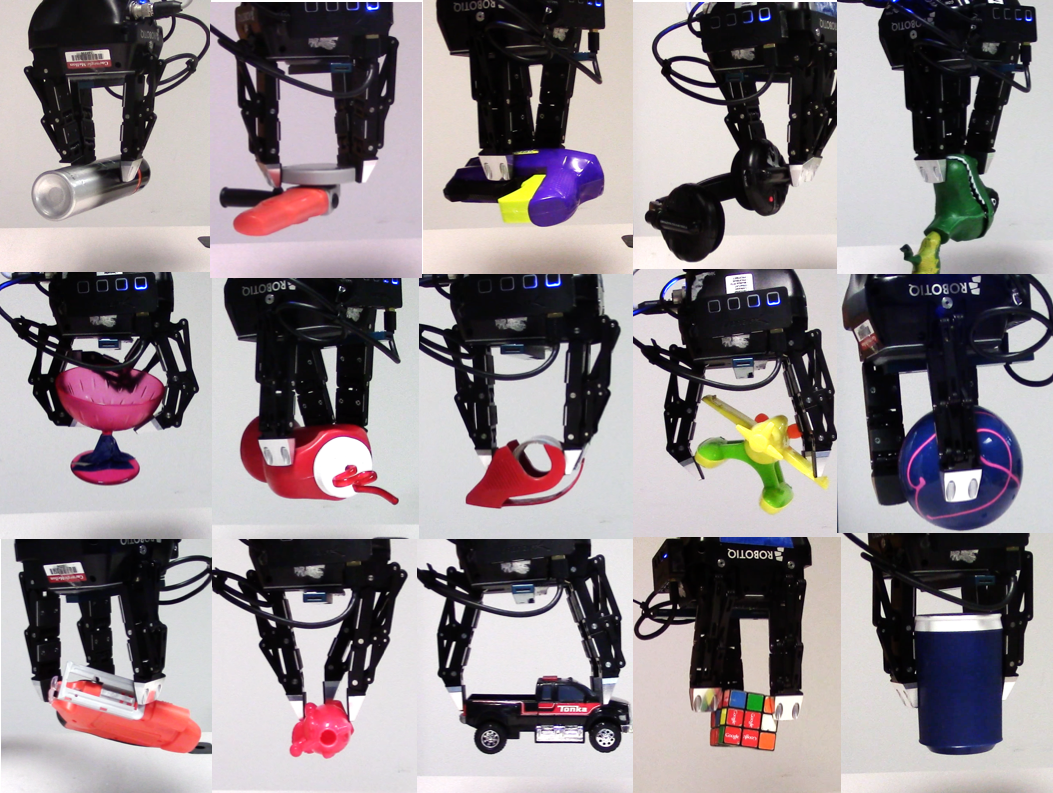}
  \caption{Some successful grasps achieved by model trained with CASSL.}
\label{fg:successful_grasps}
\end{figure}

{\bf Experimental Settings:} To quantitatively evaluate the performance of our framework, we physically tested the learned models on a set of diverse objects and measured their grasp accuracy averaged over a large number of trials. We have three test sets (shown in Fig \ref{fg:test_objects}): 1) Set A containing 10 objects seen by the robot during training 2) Set B containing 10 novel objects and 3) Set C with 20 novel objects. For Sets A and B, 5 grasps were attempted for each object placed in various random initial configurations and the results are detailed in Table \ref{tb:test}. CL0 in Table \ref{tb:test} refers to the model that was trained on the 1960 grasps collected for sensitivity analysis. Fig \ref{fg:successful_grasps} shows some of the successful grasps executed with the robot using the final model trained with CASSL (i.e. CL6). Given the long physical testing time on the largest test set C, we took the best performing model and baselines on test sets A and B and tested them on Set C. As summarized in Table \ref{tb:test}, the values reported for each model were averaged for a total of 160 physical grasping trials (8 per object). When testing, the object was placed in 8 canonical orientations (NSWE,NE,SE,SW and NW) with respect to the same reference orientation.

\begin{figure}[t]
\centering
        \includegraphics[width =3.0in]{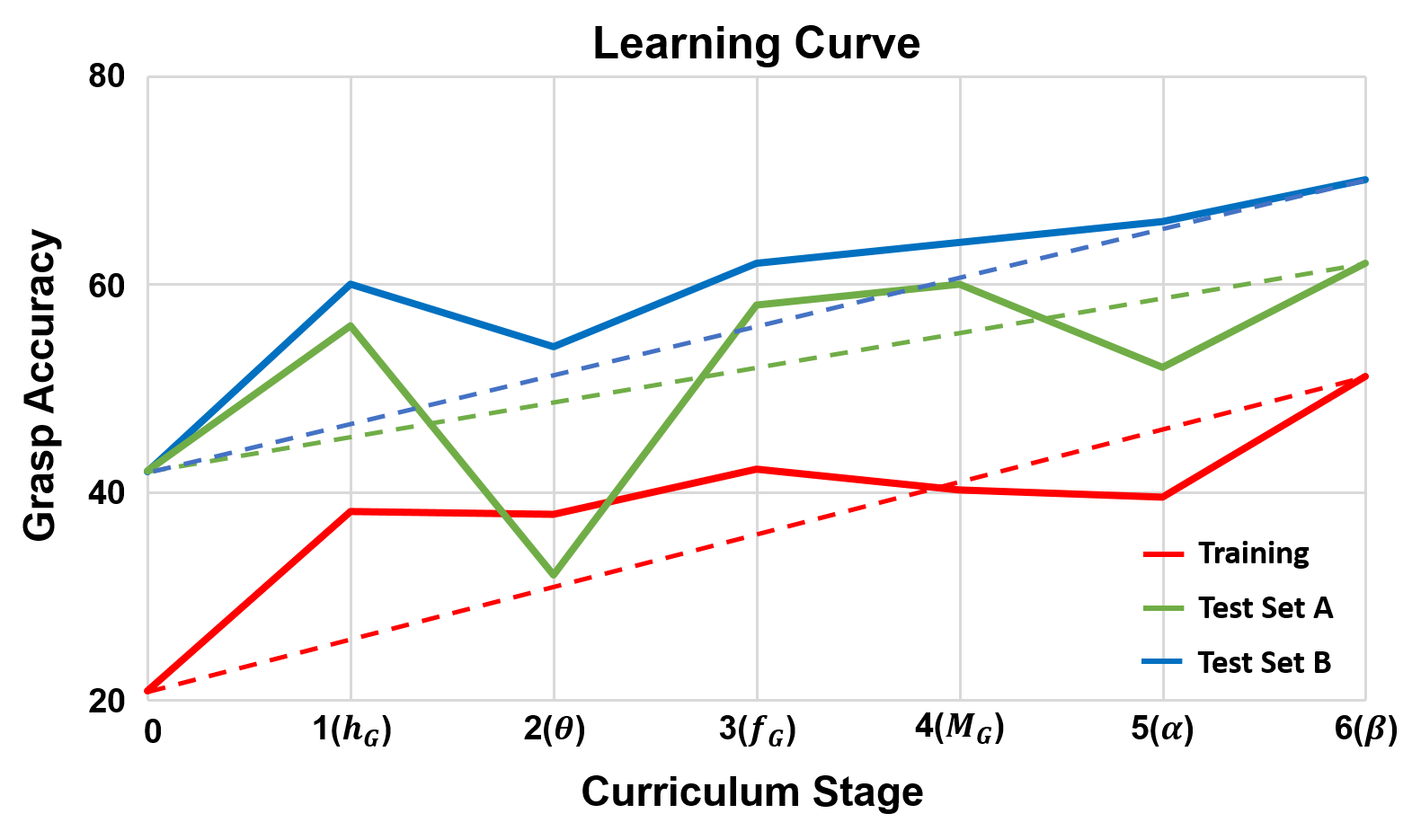}
  \caption{Variation in grasp accuracy with respect to stages in learning}
\label{fg:GraspStageAccu}
\end{figure}

{\bf Curriculum Progress:} The grasp accuracy increases with each stage of curriculum learning on Set A and B, as shown in Fig \ref{fg:GraspStageAccu}. Starting with CL0 at 41.67\%, the accuracy topped 70.0\% on Set A (Seen objects) and 62\% on Set B (Novel objects) at the end of the curriculum for the CL6 model. Note that at each stage of the curriculum, the model trained on the previous stage was used to collect around 460-480 grasps as explained in Algorithm \ref{alg:curr_train}. As expected, the performance of the models on Set A with seen objects was better than that of the novel objects in Set B. Yet, the strong grasping performance on unseen objects suggests that the CNN was able to learn a generalized visual representation to scale its inference to novel objects. There was a dip in accuracy for CL2, possibility owing to over-fitting on one of the control dimensions, but the performance recovered in subsequent stages since the models are trained with all the aggregated data.

\begin{table}[]
\centering
\caption{Results on test set with seen and novel objects}
\label{tb:test}
\begin{tabular}{|c|c|c|c|c|}
\hline
 & \multirow{2}{*}{Training} & \multicolumn{3}{c|}{Testing}   \\ \cline{3-5}
 &  & \multicolumn{1}{c|}{Set A} & \multicolumn{1}{c|}{Set B} & \multicolumn{1}{c|}{Set C} \\ \hline
CL0                     & 20.9           & 42.0           & 42.0             & -     \\ \hline
CASSL(Ours) - CL6($\beta$) & 51.1        & \textbf{70.0}& \textbf{62.0}  & 66.9 \\ \hline
CASSL(Random1)          & 42.7           & 56.0           & 54.0             & 55.6 \\ \hline
CASSL(Random2)          & 37.1           & 54.0           & 50.0             & - \\ \hline
Staged Learning \cite{pinto2016supersizing,levine2016learning} & 26.85           & 66.0          & 54.0 & 56.9 \\ \hline
Random Exploration              & 25.8           & 48.0           & 48.0             & - \\ \hline
\end{tabular}
\end{table}

{\bf Baseline Comparison:} We evaluated against four baselines, all of which are provided equal or more data than that given to CASSL. 1) \textit{Random Exploration} - Training the network from scratch with 4756 random grasps. 2) \textit{Staged Learning} \cite{pinto2016supersizing, levine2016learning} - We first trained the network with data from sensitivity analysis (i.e. CL0) and used this learned policy to sample the next 2796 grasp data points, as done in prior work. The policy was then fine-tuned with the aggregated data (4756 examples). In the third and final stage, 350 new grasp data points were sampled. This staged baseline was the training methodology used in prior work \cite{pinto2016supersizing, levine2016learning}. 3) \textit{CASSL (Random 1 \& 2)} - Instead of using sensitivity analysis to define the curriculum, two sets of randomly ranked control parameters were trained with CASSL and the performance of the final trained models is reported in Table \ref{tb:test}. The ordering for Random 1 and 2 is $[M_G, \alpha, \theta, \beta, f_G, h_G]$ (in decreasing order of priority) and $[\beta, f_G, \alpha, h_G, M_G, \theta]$ respectively. In addition to the baselines above, the CL0 model achieves a grasping rate of around 20.86\% and this could be roughly considered as the performance of random grasping trained with 1960 datapoints.

All the curriculum models (except CL0, CL2) outperformed the random exploration baseline's accuracy of 48\%. On the Set B (novel objects), CL6 showed a marked increase of 14\%, 8\% and 12\% vis-\`{a}-vis the random exploration, staged learning and CASSL (Random 2) baselines respectively. For the results on the larger Set C, CL6 still outperformed staged learning by about 10\% and CASSL (Random 1) by 11.3\%. The curriculum optimized with sensitivity analysis outperformed the random curriculum, illustrating the importance of choosing the right curriculum ranking, the lack of which can hamper learning performance.


\section{Conclusion and Future Work}
We introduce Curriculum Accelerated Self-Supervised Learning (CASSL) for high-level, high-dimensional control in this work. In general, using random sampling or staged learning is not optimal. Instead, we utilize sensitivity analysis to compute the curriculum ranking in a data-driven fashion and assign the priority for learning each control parameter. We demonstrate effectiveness of CASSL on adaptive, 3-fingered grasping. On novel test objects, CASSL outperformed baseline random sampling by 14\%, on-policy sampling by 8\% and a random curriculum baseline by 12\%. In future work, we hope to explore the following: 1) Modify the existing framework to include dynamically changing curriculum instead of a pre-computed stationary ordering 2) Investigate applications in hierarchical reinforcement learning, where high-level policy trained with CASSL is used alongside a low-level controller 3) Scale CASSL for learning in high dimensional manipulation tasks such as in-hand manipulation.

\section*{ACKNOWLEDGEMENTS}
This work was supported by ONR MURI N000141612007, NSF IIS-1320083 and Google Focused Award. Abhinav Gupta was supported in part by a Sloan Research Fellowship and Adithya was partly supported by a Uber Fellowship. The authors would also like to thank Alex Spitzer, Wen Sun, Nadine Chang, and Tanmay Shankar for discussions; Chuck Whittaker, Eric Relson and Christine Downey for administrative and hardware support.

\bibliographystyle{IEEEtran}
\bibliography{IEEEabrv,references}










\end{document}